\documentclass[letterpaper, 10 pt, conference]{ieeeconf}  

\IEEEoverridecommandlockouts                              

\usepackage{gensymb}
\usepackage{graphicx} 
\usepackage{times} 
\usepackage{amsmath} 
\usepackage{amssymb}  
\usepackage{url}
\usepackage{cite}
\usepackage{enumerate}
\usepackage{bm}
\usepackage{cases}
\usepackage{mathtools}

\usepackage{multirow}
\usepackage{colortbl}
\usepackage{booktabs} 
\usepackage{makecell}
\usepackage{algpseudocode}

\usepackage[caption=false, font=footnotesize]{subfig}
\usepackage{algorithmicx}
\usepackage{algorithm}
\usepackage{algpseudocode}
\usepackage{booktabs, ragged2e}
\usepackage{tabularx}
\usepackage{xcolor}
\usepackage[para]{threeparttable}
\usepackage{colortbl}
\usepackage{flushend} 
\usepackage[normalem]{ulem}
\useunder{\uline}{\ul}{}
\usepackage{siunitx}
\usepackage{float}
\usepackage{afterpage}

\makeatletter
\let\NAT@parse\undefined
\makeatother
\usepackage[hidelinks]{hyperref} 

\newcolumntype{C}{>{\Centering\arraybackslash}X}
\newcolumntype{L}{>{\raggedright\arraybackslash}X}
\newcolumntype{R}{>{\raggedleft\arraybackslash}X}

\begin{document}
\title{\LARGE \bf
Improving Autonomous Driving Safety with POP: A Framework for Accurate Partially Observed Trajectory Predictions

\author{Sheng Wang, Yingbing Chen, Jie Cheng, Xiaodong Mei, Ren Xin, Yongkang Song and Ming Liu}


\thanks{This work was supported by Lotus Technology Ltd. through The Hong Kong University of Science and Technology (GZ) under Cooperation Project R00082. (Corresponding author: Ming Liu.)}

\thanks{Sheng Wang, Yingbing Chen and Ren Xin are with Robotics and Autonomous Systems, Division of Emerging Interdisciplinary Areas (EMIA) under Interdisciplinary Programs Office (IPO), The Hong Kong University of Science and Technology, Hong Kong SAR, China. \texttt{\{swangei, ychengz, rxin\}@connect.ust.hk}} 

\thanks{Jie Cheng is with Electronic and Computer Engineering, The Hong Kong University of Science and Technology, Hong Kong SAR, China. \texttt{jchengai@connect.ust.hk}}

\thanks{Xiaodong Mei is with Computer Science, The Hong Kong University of Science and Technology, Hong Kong SAR, China. \texttt{xmeiab@connect.ust.hk}}

\thanks{Ming Liu is with Robotics and Autonomous Systems Thrust, The Hong Kong University of Science and Technology (Guangzhou), Guangzhou 511400, China, and also with the HKUST Shenzhen-Hong Kong Collaborative Innovation Research Institute, Futian, Shenzhen 518055, China. \texttt{eelium@hkust-gz.edu.cn}} 

\thanks{Yongkang Song is with Lotus Technology Ltd, China. \texttt{yongkang.song@lotuscars.com.cn}}


}
\maketitle

\begin{abstract}

Accurate trajectory prediction is crucial for safe and efficient autonomous driving, but handling partial observations presents significant challenges. To address this, we propose a novel trajectory prediction framework called Partial Observations Prediction (POP) for congested urban road scenarios. The framework consists of two key stages: self-supervised learning (SSL) and feature distillation. POP first employs SLL to help the model learn to reconstruct history representations, and then utilizes feature distillation as the fine-tuning task to transfer knowledge from the teacher model, which has been pre-trained with complete observations, to the student model, which has only few observations. POP achieves comparable results to top-performing methods in open-loop experiments and outperforms the baseline method in closed-loop simulations, including safety metrics. Qualitative results illustrate the superiority of POP in providing reasonable and safe trajectory predictions. Demo videos and code are available at
\href{https://chantsss.github.io/POP/}{\color{blue}{https://chantsss.github.io/POP/}.}

\end{abstract}

\renewcommand{\arraystretch}{1.0}
\section{Introduction\label{sect:intro}}

The rapid development of autonomous vehicles has brought a myriad of challenges and opportunities to both academia and industry in recent years. One of the critical aspects of self-driving technology is vehicle trajectory prediction, which provides valuable information for autonomous vehicles to assess potential risks and make informed decisions in dynamic traffic situations. Challenges in this domain include the dynamic and unpredictable nature of traffic, interactions between road users, diverse driving behaviors, sensor occlusions and limitations. Recently, data-driven approaches exhibited promising performance in prediction accuracy on challenges \cite{Argoverse}\cite{Argoverse2}. 
A motion forecasting model typically collects comprehensive information from perception signals and highdefinition (HD) maps, such as traffic light states, motion
history of agents, and the road graph. The most state-of-the-art prediction models adopt Transformers \cite{gameformer}\cite{qcnet}\cite{hivt} or graph neural networks (GNNs) \cite{tnt}\cite{lanegcn} to encode agent-agent and agent-map interactions have achieved outstanding prediction accuracy. 
Some researchers proposed to employ Self-supervised learning (SSL) to train a network for more transferable, generalizable, and robust representation learning. For example, SSL-Lanes\cite{bhattacharyya2022ssllanes} and PreTraM\cite{xu2022pretram} demonstrated that carefully designed pretext tasks can significantly enhance performance without using extra data by learning richer features.

However, these methods focus solely on fitting an inference model on a dataset with a neural network, without considering the mismatch between the distribution of the actual noisy data from the upstream module and the clean data provided on the dataset. This mismatch is mainly due to the limitations of the sensing and tracking system equipment and algorithms in the real world, which is known as the sim-2-real problem. Some works have explored how to improve the robustness of predictors from the perspective of input noise\cite{zhang2022adversarialprediction}\cite{cao2022robustprediction}, but they have neglected a key phenomenon in practical applications, namely, domain shift due to insufficient observation data, which is common to see in autonomous driving scenarios.

\begin{figure}[tb]
        \centering
        \includegraphics[width=0.8\linewidth]{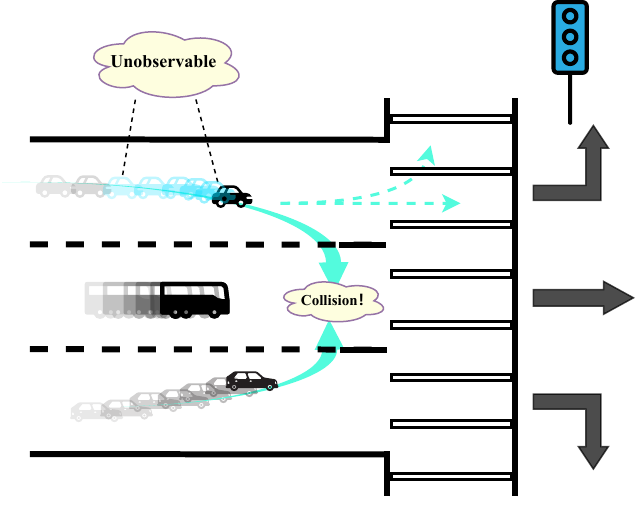}
        \caption{\textbf{Partially observed predictions in real-world situations.} In this scenario, the self-driving car is making a left turn, but another car is accidentally turning right from the left turn lane. Due to insufficient observations, the future trajectories provided by the prediction algorithm fail to include this possibility, leading to a dangerous situation.} 
        \label{fig:first_img_source}
\vspace{-2em}
\end{figure}

In concrete terms, most current learning-based prediction algorithms require a fixed-length history trajectory as output, as per the popular challenge setting. For example, Argoverse 1 (Av1)\cite{Argoverse} provides 2 seconds of history information, while Argoverse 2 (Av2)\cite{Argoverse2} requires 5 seconds of observations as input for longer-term prediction tasks. However, such algorithms developed under this fixed input length setting cannot handle real-world applications where only partial observations are available. For instance, if a previously obscured vehicle suddenly appears in the view of the ego vehicle, the predictor's inability to accurately predict its future trajectory due to insufficient observation length can lead to a collision risk, as shown in Fig. \ref{fig:first_img_source}. In later sections, we analyze upstream tracking and sensing data to shed light on this critical issue. We argue that even the most advanced architectures specialized for this task fail to process variable observation lengths. Alessio \textit{et al.}\cite{monti2022observations} propose a distillation framework that recovers a reliable proxy of the same information obtained with more input observations, but it still only supports a fixed observation length. To address this problem, we propose a new hierarchical prediction framework that can handle dynamic observation lengths by introducing an SSL mask strategy to a history reconstruction pretext task. Additionally, we employ a feature distillation scheme to transfer future extraction ability to the student model. Our contributions are summarized as follows:

\begin{itemize}
    \item Our study uncovers the critical challenge of performance degradation of trajectory predictors in the case of insufficient observations. To the best of our knowledge, this is the first comprehensive and systematic analysis of the partially observed prediction problem.

    \item We propose a flexible prediction framework, Partial Observations Prediction (POP), which employs Self-Supervised Learning (SSL) and feature distillation techniques, and which is capable of outputting stable and high-precision prediction results even when only partial observations are available.

    \item The proposed method is thoroughly evaluated on real-world datasets and a closed-loop simulator. Evaluation results demonstrate that the POP framework achieves comparable or superior performance in terms of prediction accuracy and safety metrics compared to existing state-of-the-art methods. 
    
\end{itemize}

\section{Related Work}

\subsection{Feature distillation in Motion Forecasting}

The idea of knowledge distillation is first brought up by Hinton etal. \cite{hinton2015distilling} to transfer knowledge from a large, complex model (teacher) to a smaller, simpler model (student) for model compression purpose. 
In the context of trajectory forecasting, knowledge distillation has been used to make a model immune to incorrect detection, tracking, fragmentation, and corruption of trajectory data in crowded scenes, by distilling knowledge from a teacher with longer observation to a student with much fewer observations\cite{monti2022observations}. Although the performance is nearly retained though. This is not reasonable in more challenging autonomous driving scenes when we have longer observations(e.g. 49 frames) and only pick the first 2 frames and throw the others useful information.
Inspired by \cite{selfd-istillation}, our aim is not to compress a model yet to improve its performance. This procedure is usually referred to as self-distillation, since the student network shares the same architecture of its teacher. Similarly to \cite{porrello2020robustdistill}, our approach sets up asymmetric networks: the student is encouraged to overcome its knowledge
gap by following the guide of its teacher, eventually boosting its performance. We demonstrate that knowledge
distillation can lead to effective predictions even when the
model has access to very few observations.

\subsection{SSL:Self-supervised Learning in Motion Forecasting}

Self-supervised learning (SSL) has been widely explored and utilized in various research domains\cite{WhenDoes, SSLSurvey}. SSL leverages the inherent structure or patterns present in unlabeled data to learn useful representations or tasks. In previous trajectory prediction work, SSL has shown promising capabilities in improving the prediction accuracy. PreM\cite{xu2022pretram} focuses on connecting trajectory and scene context, enhancing their representations for trajectory forecasting. They do not have the task of reconstructing history. A recent work F-MAE\cite{FMAE} proposes to mask agents' trajectories and lane segments and reconstruct masked elements using a prediction head, at last fine tune the motion forecasting task. However, its reconstruction mechanism is in terms of the whole trajectory. In contrast, in our proposed method, we take the state at each time step as the reconstruction unit, which is more consistent with the real-life POP situation. A very recent work\cite{xu2023uncovering} used a temporal decay module to estimate the missing observation, and treat the imputation as a training objective that is jointly optimized with the motion forecasting task. However, it neglects the effectiveness of reconstructing missing observations to planning task.
This is extremely vital for safe autonomous driving systems, and we will elaborate on this later in the experimental module.
\section{Problem Formulation}


We adopt a structured vectorized representation to depict the map and agents. We denote the past trajectories of agents as $X_H = \{x_i\}$, where $x_i \in \mathbb{R}^{T_H \times D}$ indicating the location, yaw angle, and velocity of agent $i$ at previous $T_H$ time steps. The road map is denoted as $L = \{l_i\}$, where $l_i \in \mathbb{R}^{N \times F}$ representing $i_{th}$ lane has $N$ segments and each segment has $F$ lane semantic attributes (e.g., intersections and crosswalks). The forecasting task aims to generate $T_F$ steps future trajectories :

\begin{equation}
    Y_F = f(X_H * M_H, L),
\end{equation}
where $M_H = \{m_i\}$, $m_i \in \mathbb{R}^{T_H \times 1}$ indicating the validity for history state at previous $T_H$ time steps. In contrast to the previous definition of trajectory prediction, we consider that the history trajectory of the focal agent does not necessarily satisfy completeness, and thus we set ${m^{T_H}_i} = 0$ for states that are not valid at $T_H$.

\section{Methodology}

\subsection{Overview}
\begin{figure*}[t]
        \centering
        \includegraphics[width=0.90\textwidth]{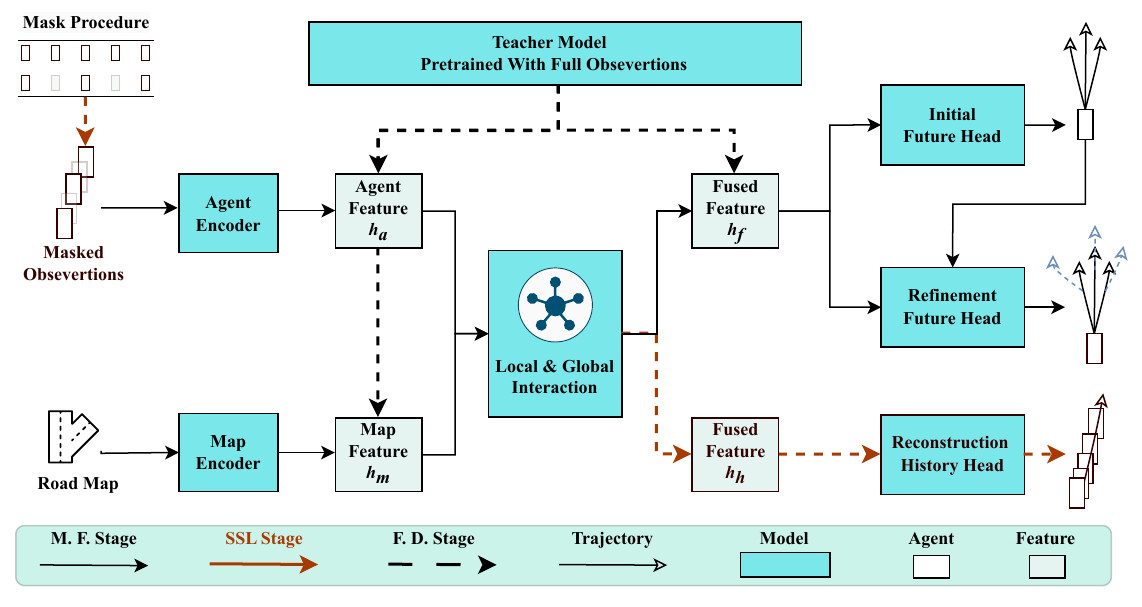}
        \caption{\textbf{Overview of POP.} Our method consists of three stages. The motion forecasting stage involves training a teacher model with complete observations. The SSL stage consists a mask procedure and a history reconstruction pre-task. During the distillation stage, the teacher model's parameters are frozen, and a feature distillation strategy is applied to the hidden features.} 
        \label{fig:overview}
\vspace{-1.5em}
\end{figure*}

The overview framework is shown in Fig. \ref{fig:overview}. Our prediction framework comprises three stages. In the first stage, we train a teacher model using complete observations. The inference stage is the standard prediction process: $h_a$ and $h_m$ features are generated using road maps and history states through an encoder consisting of a multi-layer perceptron (MLP) and a location embedding layer. A series of attention modules are used to capture the interaction information between elements, and the decoders generate the initial future guess and refinement. During the SSL stage, we enable a mask procedure and use partial observations as input, along with a history reconstruction pre-task to reconstruct missing observations. In the Distillation step, we freeze the teacher model's parameters and the feature distillation strategy is used to transfer knowledge to intermediate features of partial observations, while keeping the mask procedure on. It is worth noting that, unlike previous approaches adopted SSL that aim to improve predictor performance with complete observations, we consider the distillation task with partial observations as the final fine-tuning task.

\subsection{Motion Forecasting Stage}

In this stage We follow the inference pipeline of QCNet\cite{qcnet} and consider it as a strong baseline in experiments in the later section. We build a local spacetime coordinate system for each scene element to encode scene representations. These representations are transformed to Fourier features and passed through an MLP to obtain relative positional embeddings $R$. Factorized attention and self-attention mechanisms are applied to $\{X_H, R\}$ and $\{L, R\}$ respectively to obtain hidden encoded features $ha$ and $hm$. We adopt the architecture of detection transformer in both the initial and refinement decoder modules to address the one-to-many problem, allowing multiple learnable queries to cross-attend the scene encodings and a MLP to decode trajectories. A slight difference in refinement decoder is that a gated recurrent unit is used to embed each trajectory anchor, and we take its final hidden state as the mode query. Taking the output of the proposal module as anchors, we let the refinement module predict the offset to the proposed trajectories and estimate the likelihood of each hypothesis.






    
\subsection{Self-supervised learning Stage}
    Recall the motivation, our aim is to make predictors robust to insufficient observations. An intuitive to achieve this goal is to add noise or perform data augmentation. Since the partial observation phenomenon varies from time to time, we are not accessible to a uniform real-world noise distribution. Thus we build a pretext task from reconstructing the random masked input history. Specifically, a mask procedure and a reconstruction branch is built when performing the SSL stage. It is similar to the future prediction branch structurally. It only differs from the output dimension of decoder head, which means we use $T_H$ as the prediction horizon. 



\subsection{Feature Distillation Stage}
In order to maintain the predictive capabilities of teachers in the face of insufficient observations, our training strategy involves transferring knowledge from the entire input sequence. To achieve this, we manipulate hidden features from the encoder and interaction module. Specifically, we ensure that all of the student's features correspond to those in the teacher network. This is accomplished through a feature distillation loss, which is defined as the mean squared error (MSE) between the two feature representations:
\begin{equation} 
\mathcal{L_{D}} = \frac{1}{d}\sum\left\|(h^T - h^S)\right\|^2,
\end{equation}
where $d$ is the dimension of the corresponding feature representations, $h^S = \{h_a^S, h_m^S, h_f^S\}$, $h^T = \{h_a^T , h_m^T, h_f^T\}$. The loss encourages the student model to learn feature representations that are similar to those of the teacher model.

\subsection{Training Objectives}
Our goal is to build, in three steps, a model capable of accurately predicting future locations when only partial observations are available. To train the teacher model in the first stage, we employ negative log-likelihood loss and winner-take-all training strategy to optimize the best-predicted future trajectory. The training loss in this stage is: 
\begin{equation}
    \mathcal{L_{MF}} = \mathcal{L}_{init} + \mathcal{L}_{refine} + \alpha \mathcal{L}_{cls}, 
    \label{stage1_loss}
\end{equation}
where the classification loss is added to optimize the mixing coefficients and $\alpha$ is parameter to balance regression and classification. 

\begin{equation}
    \mathcal{L_{SSL}} = \mathcal{L_{MF}} + \beta \mathcal{L}_{recons}, 
    \label{ssl_loss}
\end{equation}
\begin{equation}
    \mathcal{L_{FD}} = \mathcal{L_{MF}} + \lambda \mathcal{L}_{D}. 
    \label{lfd_loss}
\end{equation}

When perform SSL stage or feature distillation stage, we keep the $L_{MF}$ and add a reconstruction loss or distillation loss with balance parameter $\beta$ or $\lambda$ respectively as shown in Eq. \ref{ssl_loss} and Eq. \ref{lfd_loss}.





\section{Preliminary Analysis}\label{sect:experiments}

Before diving into the experiment section, we will explore two questions to recall the motivation and further validate the effectiveness of the proposed method. \textit{First of all,  is it common to see inefficient observation situations in the real world? Secondly, are existing state-of-the-art (SOTA) trajectory prediction methods able to handle the partial observation problem?} 
\subsection{Observation Distribution Analysis}
\begin{figure}[t]
        \centering
        \includegraphics[width=0.80\linewidth]{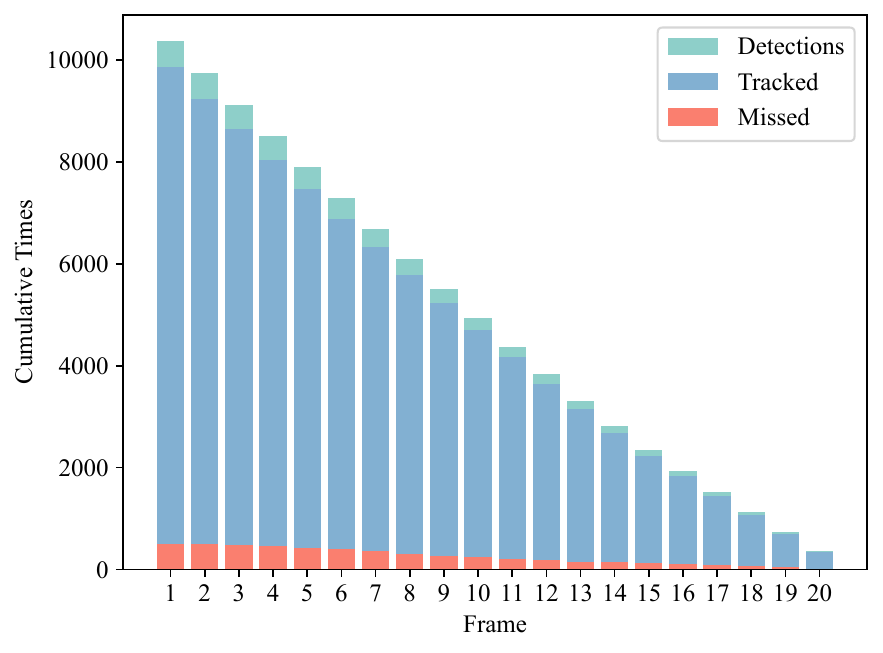}
        \caption{\textbf{Distribution of observations with standard 20 frames.} The gray bars represent the detection of each frame for a fixed observation period, while the blue bars represent tracking. The Red indicates tracking failures.} 
        \label{fig:ob_distribition}
\vspace{-2em}
\end{figure}

To answer the first question, we investigate the Av1 tracking dataset, which is a collection of 113 log segments with 3D object tracking annotations. These log segments vary in length from 15 to 30 seconds and collectively contain a total of 11,052 tracks. 
We simulate the original observations using the tracking baseline method \cite{Tracker}, which won first place on the Argoverse 3D tracking text set. According to the Av1 motion forecasting challenge standard, 20 frames are used for a complete observation. The observations are constructed frame by frame from the beginning of the appearance of the target until the target disappears. Fig. \ref{fig:ob_distribition} shows the distribution of the observation, indicating that the prediction algorithm for autonomous driving applications is often unable to satisfy the complete 20-frame observations on the dataset due to occlusion, limitations in sensing range, large speed differences between vehicles, \textit{etc}. 

\subsection{Observation Length Evaluation Analysis}
\begin{figure}[t]
        \centering
        \vspace{-0.05em}\includegraphics[width=0.78\linewidth]{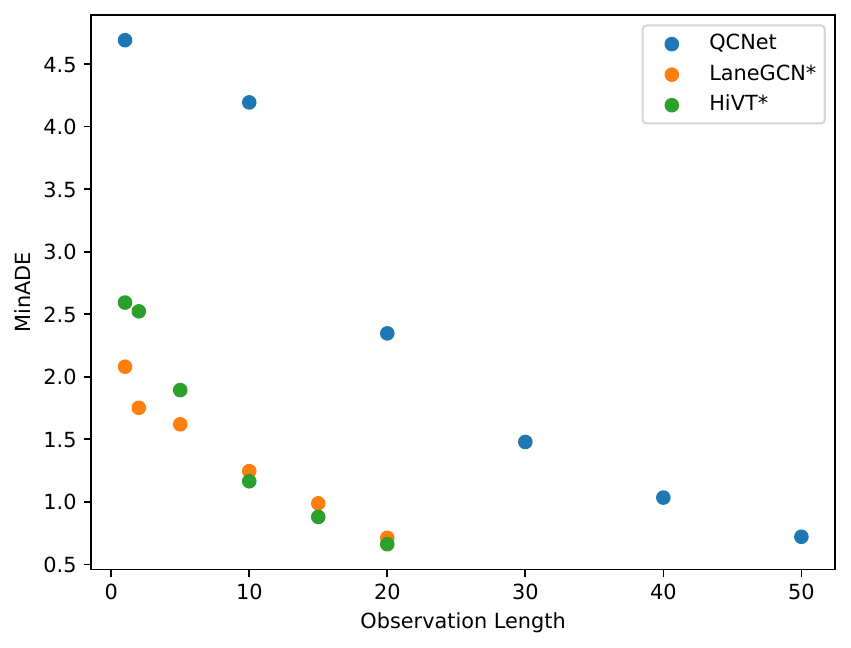}
        \caption{\textbf{Observation length evaluation analysis on Av1 and Av2.} Methods that are evaluated on Av2 are marked with symbol ``*''.} 
        \label{fig:pop_analysis}
\vspace{-1em}
\end{figure}

We evaluate the MinADE$_5$ performance of three popular predictors by given observations of varying lengths. As shown in Fig. \ref{fig:pop_analysis}, the predictor's performance is directly correlated with the length of the observations. In other words, the more information from observations used as input, the greater the accuracy of the predictions made by the predictor. This relationship is especially pronounced when dealing with long prediction horizon task. Therefore, we conclude that the currently available SOTA prediction methods are unable to effectively handle situations with partial observations.

\section{Experimental Results}\label{sect:experiments}

\subsection{Open-Loop Experiment}

\subsubsection{Dataset}
We evaluate our method on both Av1 and Av2. The former requires a 2s history input and a 3s future trajectory output, and the latter one focuses on a long term 6s prediction and a longer corresponding 5s history input.

\subsubsection{Metrics}
For the open-loop evaluation, we use the official benchmark metrics, including MinADE$_k$, MinFDE$_k$, MR$_k$, and brier-MinFDE$_k$. These metrics are calculated based on the trajectory with the closest endpoint to the ground truth over $k$ predictions.

\subsubsection{Implementation Details}

For all experiments, we utilize cosine learning rate decay with a weight decay of 0.0001 and an initial learning rate of 0.0005. The model is trained on 4 RTX3090s, and the pre-trained teacher model is obtained from the officially released checkpoint\cite{qcnet}. The hyperparameters $\alpha$, $\beta$, $\lambda$ are set to 1, 0.5, 0.5 respectively. As road structures and traffic rules vary across regions, a generalized observation distribution cannot be obtained. We adopt a random drop scheme during training. No data augmentation or model ensemble techniques are employed.


\subsubsection{Comparison with State-of-the-Art}
We compare our method to the SOTA without ensembles on the Av2 test set in the single-agent setting, and the results are reported in Table \ref{table:SOTA} (upper group). Our method achieves results comparable to the best-performing methods, ranking third on the MinADE$_6$ metric. Our performance is also comparable to that of F-MAE, which uses the SSL strategy with complete observations. In later experiments, we will demonstrate the superiority of POP under partial observation conditions. To evaluate the generalization of our method, we conduct experiments on the Av1 test set using HiVT as the backbone. Although our POP-H performance is slightly degraded compared to HiVT, this is due to the use of incomplete observations to train the predictor, which increases the difficulty of fitting the network. Nonetheless, our performance is still comparable to SOTA methods, such as LTP and mmTrans.

\begin{table}[h]
\caption{Closed-loop Simulation Results on Commonroad Dataset.}
\centering
\vspace{-0.5em}
\begin{threeparttable} 
\begin{tabularx}{\linewidth}{p{9.5mm}p{5.0mm}|CCCCC}
\Xhline{2\arrayrulewidth}
 Method & P-K & DIST$\uparrow$  & JERK$\downarrow$ & RC$\downarrow$ & CT$\downarrow$ & RCT$\downarrow$ \\ \hline \hline
 HiVT & 1 & 30.19  & 3.23 & 0.657 & 7 & 1 \\ 
 HiVT & 3 & 28.80 & \textbf{2.87} & 0.636 & 2 & 1 \\ 
 POP-H & 1 & \textbf{30.97}  & 4.07 & 0.641 & 5 & \textbf{0} \\
 POP-H & 3 & 28.38  & 3.07 & \textbf{0.616} & \textbf{1} & 1 \\ \hline \hline
 HiVT & 1 & 40.77  & \textbf{3.24} & 0.608 & 21 & 5 \\ 
 POP-H & 1 & \textbf{41.35}  & 3.50 & \textbf{0.576} & \textbf{17} & \textbf{4} \\

\Xhline{2\arrayrulewidth}
\end{tabularx}

\smallskip
\scriptsize
\begin{tablenotes}
\item The top group records performance in 91 highly interactive scenarios, while the bottom group represents that in all.  \\
\end{tablenotes}
\end{threeparttable}
\label{table:multi_modal_experiments}
\vspace{-2.5em}
\end{table}

\begin{table}[h]
\centering
\caption{Open-loop Simulation Results on Commonroad Dataset.}
\centering
\vspace{-1em}
\begin{tabularx}{\linewidth}{lCCCC}
\toprule
         Model & MinADE$_1$  & MinFDE$_1$  & RMinADE$_1$ & RMinFDE$_1$ \\
\midrule

        HiVT &     \textbf{2.51} &    \textbf{5.21}  &     3.22   &   8.18     \\
        POP-H &   2.54 &    6.23  &      \textbf{2.66} &    \textbf{6.30}    \\ 
\bottomrule
\end{tabularx}

\smallskip
\scriptsize
\begin{tablenotes}
\item RMinADE$_1$, RMinFDE$_1$ are collected with randomly observation lengths. \\

\end{tablenotes}
\label{table:Open-loop Results on Interaction Dataset}
\vspace{-1.5em}
\end{table}

%
%
%

\subsubsection{Ablation Study}
 The ablation study results are presented in Table \ref{tab:ablation}. It shows that QCNet performs poorly with partial observations. In particular, when fewer than 30 frames are available, the prediction error increases to two times compared to when the full observations are input. However, when using either the SLL or Distillation strategies individually, the performance shows a notable improvement in the POP case. This highlights the effectiveness of both stages in our design. While the SLL-only strategy achieves the same level of performance as the POP strategy with complete observations, POP outperforms SLL when there are less than 50 observations. Our findings suggest that feature distillation, treated as a fine-tuning task after SLL, allows for further feature learning and better performance.

\subsection{Closed-Loop Experiment}

\subsubsection{Simulation Setting}

The closed-loop experiments are conducted in the interactive scenarios from CommonRoad \cite{CommonRoad}, with a simulation setup similar to \cite{IR-STP}. Considering the limited field-of-view and perception range of the autonomous vehicle (AV) in real-world scenarios, we limit our predictions to neighboring vehicles within a 50m radius. At each step, we use POP-H or HiVT to predict the future trajectories of agents over a 6s horizon with a 0.5s interval. The planning process for collision check directly incorporates the $K$ most probable prediction outcomes for each agent. The prediction model is pre-trained using feature data extracted from the closed-loop simulation itself. Each simulated scenario incorporates a designated task route to evaluate the planning performance, as shown in Fig. \ref{fig:demo}. The collision avoidance planner guides the AV along the provided task route. If the implemented algorithm fails to identify a valid solution, the AV executes a stop behavior along the generated path with a deceleration of $-4.0\,m/s^2$. All other agents are controlled by the intelligent driver model.

\subsubsection{Metrics}

We adopt the following metrics: DIST: Average completion distance of the AV along the given route in each scenario. JERK: Average jerk cost reflecting the planned trajectory's smoothness. RC: Reaction cost of other traffic agents, defined as the average deceleration efforts of nearby agents within a 40-meter range. CT: Total number of valid collision times experienced by the AV, excluding collisions at the rear and collisions with agents when the AV is stationary. RCT: Collision times at the rear of the AV.

\subsubsection{Quantitive Results}
We present the open-loop prediction performance of HiVT and POP-H in Table \ref{table:Open-loop Results on Interaction Dataset}, focusing on MinADE$_1$ and MinFDE$_1$ since we use the most likely prediction trajectory for collision checking in the closed-loop simulation. Results show that vanilla HiVT's performance significantly drops with random observations, while POP-H remains stable under both complete and incomplete observations due to our design for handling incomplete observations during training. The closed-loop results are reported in Table \ref{table:multi_modal_experiments}. For each metric, the best result is in bold. As shown in the bottom group, our proposed method outperforms HiVT in almost all metrics, particularly in safety, with a 25\% reduction in collisions. HiVT yields slightly better jerk results but at the cost of less driving distance and a higher number of collisions. POP-H achieves favorable performance in 91 high interactivity scenarios, particularly when using 3 predicted trajectories for collision detection. What's more, POP-H reduces collisions to 1 and achieves the lowest RC, facilitating friendly driving to other vehicles.

\subsubsection{Qualitative Results}

\begin{figure*}[htbp]
        \centering
        \includegraphics[width=1\textwidth,height=0.5\textwidth]{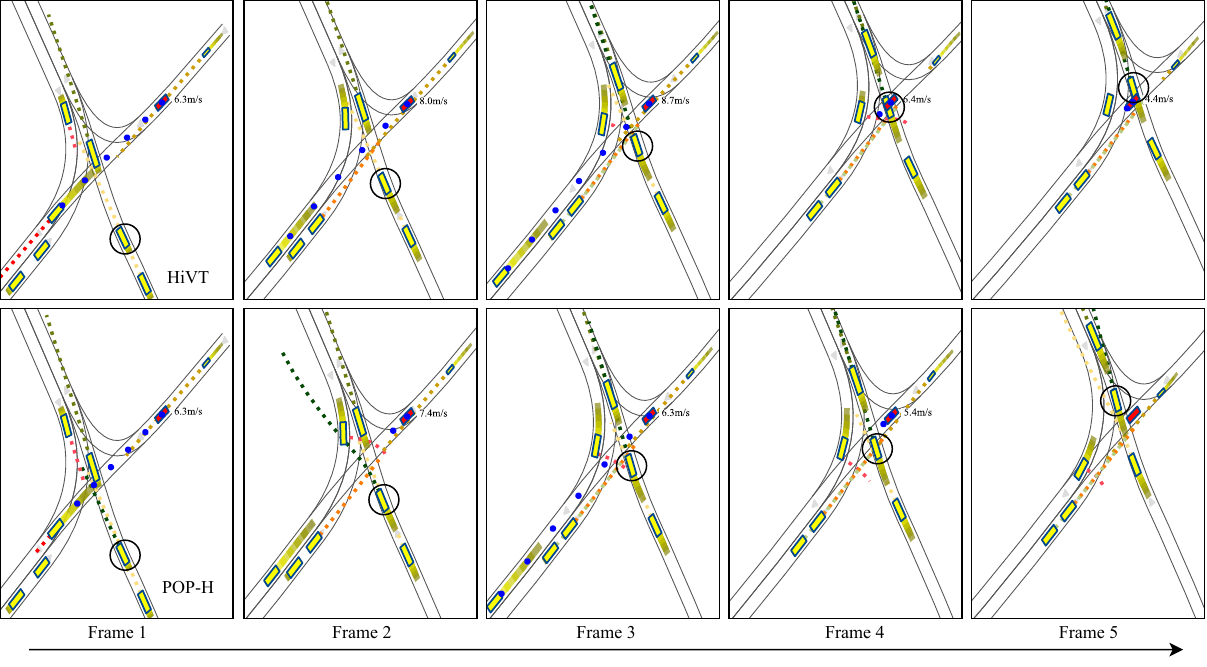}
        \vspace{-2em}
        \caption{\textbf{A collision scenario caused by limited observations.} The AV (red) is surrounded by yellow vehicles, the blue scatter line is the planned trajectory of the AV, and the predictions of the AV for other vehicles are marked with scatter lines of other colors. Due to limited observations, the HiVT predictor inaccurately predicted the future trajectory of a vehicle (black circle), causing the AV to accelerate and ultimately collide. In contrast, the AV equipped with the POP-H predictor exhibited superior predictions, ensuring safety.} 
        \label{fig:demo}
        \vspace{-1em}
\end{figure*}

We demonstrate a simulation scenario where the AV navigates through a congested traffic intersection, as shown in Figure \ref{fig:demo}. During the initial phase of the simulation, the AV intends to traverse the intersection with a planned speed of 6.3 m/s. However, due to insufficient observation, the HiVT predictor inaccurately predicts the future trajectory in the first two frames. As a result, the AV fails to account for the movement of the vehicle below and begins to accelerate. By frame 3, the speed has already reached 8.7 m/s, making it too late to decelerate and leading to a collision. In contrast, POP-H consistently provides more reasonable predictions (indicated by the black scatter line) from frame 1 to frame 5, ensuring a higher level of safety.


\begin{table*}[t]
\centering
\vspace{1em}
\caption{Comparison with State-of-the-Art Methods on Argoverse test set.}
\centering
\begin{tabularx}{\linewidth}{lCCCCCCC}
\toprule
         Model & MinADE$_6$ & MinADE$_1$ & MinFDE$_6$ & MinFDE$_1$ & brier-MinFDE$_6$ & MR$_1$ & MR$_6$\\
\midrule

        GANet\cite{GAnet} &     0.72 &    1.77  &     1.34 &     4.48 &  1.96   &     0.17 &  0.59        \\
      MTR\cite{MTR} &      0.73&     1.74 &     1.44 &     4.39 &  1.98    &     0.15 &  0.58       \\
     GoRela\cite{cui2022gorela} &     0.76  &  1.82  &     1.48 &    4.62 &  2.01    &     0.22 &  0.66    \\
        F-MAE\cite{FMAE} &      0.71 &    1.74 &     1.39 &     4.36 &  2.03     &     0.17 &  0.61     \\
        QCNet\cite{qcnet} &     0.65 &    1.69  &     1.29 &     4.30 &  1.91     &     0.16 &  0.59      \\
        POP-Q (ours) &      0.72 &    1.86  &     1.46 &   4.84  & 2.08  &     0.20 &  0.61    \\ \hline \hline

   LaneGCN\cite{lanegcn} &      0.87&     1.71&     1.36 &     3.78 &  2.05    &     0.59 &  0.16    \\
     mmTrans\cite{mmtrans} &      0.84&     1.77  &    1.34  &     4.00 &  2.03    &     0.61 &  0.15    \\
      LTP\cite{LTP} &     0.83 &    1.62  &     1.30 &      3.55&  1.86      &     0.56 &  0.15     \\
      HiVT\cite{hivt} &    0.77 &     1.60 &     1.17 &     3.53 &  1.84       &     0.55 &  0.13    \\
        ADAPT\cite{ADAPT} &     0.79 &     1.59 &     1.17 &     3.50 &  1.80        &     0.54 &  0.12   \\
        POP-H (ours) &     0.83 &     1.73 &     1.32 &     3.83 &  1.99        &     0.59 &  0.15   \\
\bottomrule
\end{tabularx}
\label{table:SOTA}
\vspace{-0.5em}
\end{table*}

\begin{table*}[htbp]
  \centering
  \caption{Ablation study of MinADE/MinFDE among different training strategies on Argoverse validation set.}
    \begin{tabularx}{\textwidth}{ccc|*{6}{>{\centering\arraybackslash}X}c} 
    \toprule
    &Training strategy &  &\multicolumn{6}{c}{MinADE$_6$/MinFDE$_6$} \\
\cmidrule{1-3}     
\cmidrule{4-10} 
Scratch  &SSL  &Distill.  & Obs. = 1 & Obs. = 10 & Obs. = 20 & Obs. = 30 & Obs. = 40  & Obs. = 50& Obs. = random\\
    \midrule
    \checkmark  & & & \underline{4.69}/\underline{8.79} & \underline{4.28}/\underline{7.52}  & \underline{2.48}/\underline{4.78} & \underline{1.54}/\underline{3.13} & \underline{1.05}/\underline{2.08} & \textbf{0.72/1.25} & \underline{2.37}/\underline{4.43}\\
    \checkmark & \checkmark & & 4.03/7.45 & 2.31/4.27 & \textbf{1.10/2.05} & 0.93/1.72 & 0.85/1.57 & 0.79/1.45 & 1.47/2.72 \\
    \checkmark & & \checkmark & 4.04/7.48 & 2.37/4.37 & 1.19/2.25 & 1.00/1.88 & 0.93/1.75 & \underline{0.88}/\underline{1.66} & 1.52/2.83 \\
    \checkmark &\checkmark &\checkmark  & \textbf{4.03/7.44} & \textbf{2.24/4.16} & 1.12/2.08 & \textbf{0.92/1.70} & \textbf{0.84/1.54} & 0.79/1.45 & \textbf{1.42/2.61}\\
    \bottomrule
    \end{tabularx}

\smallskip
\scriptsize
\begin{tablenotes}
\RaggedRight

\end{tablenotes}
\label{tab:ablation}
\vspace{-1em}
\end{table*}


\section{CONCLUSIONS \label{sect:conclusion}}

We present a novel prediction framework called POP, which first employs self-supervised pre-training to help the model learn to reconstruct history, and then utilizes feature distillation as a fine-tuning task to transfer knowledge from the teacher model, which has been pre-trained with complete observations, to the student model, which has only few observations. The experiments show that compared with the existing state-of-the-art predictors, our method is able to achieve high and stable open-loop prediction accuracy both in the case of complete observations and few observations. Moreover, our method significantly enhances the safety of the autonomous driving system in the closed-loop simulation. One possible future effort is to consider exploring which observations would introduce serious hazards.



\bibliographystyle{IEEEtran}
\bibliography{IEEEabrv,ref}

\end{document}